\documentclass[letterpaper, 10 pt, conference]{ieeeconf}  % Comment this line out if you need a4paper

\IEEEoverridecommandlockouts                              % This command is only needed if 
\usepackage{graphics} % for pdf, bitmapped graphics files
\usepackage{epsfig} % for postscript graphics files
\usepackage{mathptmx} % assumes new font selection scheme installed
\usepackage{times} % assumes new font selection scheme installed
\usepackage{amsmath} % assumes amsmath package installed
\usepackage{amssymb}  % assumes amsmath package installed
\usepackage{amsfonts}
\usepackage{tikz}
\usepackage{pgfplots}
\usepackage{paralist}

\usepackage{todo}

\title{\LARGE \bf
Congestion and Scalability in Robot Swarms: a Study on Collective Decision Making
}

\author{Karthik Soma$^{1}$, Vivek Shankar Vardharajan$^1$, Heiko Hamann$^{2}$ and Giovanni Beltrame$^{1}$% <-this % stops a space
\thanks{*This work was supported by the Natural Science and Engineering Council
  of Canada (NSERC).}% <-this % stops a space
\thanks{$^{1}$Karthik Soma, Vivek Shankar Vardharajan and Giovanni Beltrame are with École Polytechnique de Montréal, 2900 Boul Édouard-
Montpetit, Québec, CA {\tt\small {karthik.soma}@polymtl.ca}}%
\thanks{$^{2}$ Heiko Hamann is with the Department of Computer and Information Science, University of Konstanz,
78457, Konstanz, Germany.%
}%
}

\begin{document}

\maketitle
\thispagestyle{empty}
\pagestyle{empty}

%%%%%%%%%%%%%%%%%%%%%%%%%%%%%%%%%%%%%%%%%%%%%%%%%%%%%%%%%%%%%%%%%%%%%%%%%%%%%%%%
\begin{abstract}
  One of the most important promises of decentralized systems is scalability,
  which is often assumed to be present in robot swarm systems without being
  contested. Simple limitations, such as movement congestion and communication
  conflicts, can drastically affect scalability. In this work, we study the
  effects of congestion in a binary collective decision-making task. We
  evaluate the impact of two types of congestion (communication and movement)
  when using three different techniques for the task: Honey Bee inspired,
  Stigmergy based, and Division of Labor. We deploy up to 150 robots in a
  physics-based simulator performing a sampling mission in an arena with
  variable levels of robot density, applying the three techniques. Our results
  suggest that applying Division of Labor coupled with versioned local
  communication helps to scale the system by minimizing congestion.
%is minimizing congestion the same as decongesting? 
%also I am not sure if is one decision variable or two decision variables. 
%multi hop communication is also important I think
\end{abstract}

%%%%%%%%%%%%%%%%%%%%%%%%%%%%%%%%%%%%%%%%%%%%%%%%%%%%%%%%%%%%%%%%%%%%%%%%%%%%%%%%
\section{INTRODUCTION}
Swarm robotics takes inspiration from natural swarms to design coordinated
behaviors. Since natural swarms exhibit properties like scalability, fault
tolerance, robustness, and parallelism, it is often assumed that these would
also be present in artificial systems like robot swarms~\cite{Brambilla2013}.
Designing robot swarms with local control rules to attain a global swarm
behavior through emergence alone might not be sufficient to ensure scalability.
Practical constraints such as crowding and communication issues hinder the
scalability of these systems and affect the deployment of robot swarms in
real-world scenarios~\cite{DorigoReflections}. In general, when robots in a
swarm share access to a resource (whether a communication medium or physical
space), it often gives rise to congestion.

%local communication and coordination strategies?
Consequently, designing and deploying robot swarms involves choosing local
communication and coordination strategies and adapting to a swarm size that will
limit congestion. Making the swarm size too large could conversely affect the
task performance, giving rise to an optimal swarm size to maximize
performance~\cite{superlinear_heiko_2018}. In some application scenarios, the
swarm size could not be chosen, and the system must perform reasonably even
when congested. We believe it is fundamental to understand the role of
congestion to address and design strategies to achieve optimal performance for
robot swarms.
\begin{figure}[t!]\label{fig:cartoon}
    
    \begin{tikzpicture}
        \begin{axis}[
            width = 1.15\columnwidth,
            height = 7cm,
            xmin=0,
            xmax=100,
            ymin=0,
            ymax=100,	
            grid=none,
            axis line style={draw=none},
            tick style={draw=none},
            xticklabel=\empty,
            yticklabel=\empty,
            ]
        \node 
        at (axis cs:50,95) 
        (Virtual Stigmergy) {\small \textcolor{blue}{\textbf{Virtual Stigmergy}}};
    
    \node 
        at (axis cs:50,50) 
        (Nest) {\small \textcolor{blue}{\textbf{NEST}}};

    \node 
        at (axis cs:15,50) 
        (black) {\small \textcolor{blue}{\textbf{ZONE B}}};
    
    \node 
        at (axis cs:85,50) 
        (white) {\small \textcolor{blue}{\textbf{ZONE A}}};
    
    % \node 
    %     at (axis cs:75,145)
    %     (comm) {\small \textcolor{black}{Communication Conflict}};
    
    % \node 
    %     at (axis cs:25,145)
    %     (Movement) {\small \textcolor{black}{Movement Congestion}};
    
    % \node 
    %     at (axis cs:50,86) 
    %     (black board) {\tiny \textcolor{black}{Distributed shared table}};

\draw[blue, very thick] (0,0) rectangle (100,100);
\draw[black, very thick] (27,0) rectangle (73,100);
\draw[orange, very thick] (35,65) rectangle (65,90);
% \draw[orange, very thick] (55,105) rectangle (70,140);
% \draw[green, very thick] (90,130) circle (0.6cm);
% \draw[green, very thick] (90,112) circle (0.6cm);

% \draw (50,105) --  (50,150);
% \draw[->] (90,130) --  (67.5,127.5);
% \draw[->] (90,112) --  (67.5,117.5);
% \draw[->] (15,125) --  (25,125) node[midway,above] {\small $\overrightarrow{v}$};
% \draw[->] (35,125) --  (25,125) node[midway,above] {\small $\overrightarrow{v}$};

\draw[->] (Nest.north east) -- (white.north west) node[midway,above] {\tiny Phototaxis};
\draw[->] (Nest.north west) -- (black.north east) node[midway,above] {\tiny Antiphototaxis};
\draw[->] (white.south west) -- (Nest.south east) node[midway,below] {\tiny Antiphototaxis};
\draw[->] (black.south east) -- (Nest.south west) node[midway,below] {\tiny Phototaxis};

\node at (axis cs:85,89) (nd10) {\includegraphics[scale=0.05,width=1cm]
                {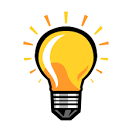}};

\node at (axis cs:50,12) (nd11) {\includegraphics[scale=0.15,width=1cm]
                {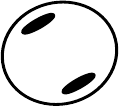}};

\node at (axis cs:40,25) (nd12) {\includegraphics[scale=0.15,width=1cm]
                {figures/khepera.pdf}};

\node at (axis cs:60,25) (nd13) {\includegraphics[scale=0.15,width=1cm]
                {figures/khepera.pdf}};

\node at (axis cs:60,75) (nd14) {\includegraphics[scale=0.1,width=0.5cm]
                {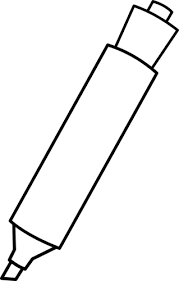}};

% \node at (axis cs:15,125) (nd15) {\includegraphics[scale=0.1,width=0.7cm]
%                 {figures/khepera.pdf}};

% \node at (axis cs:35,125) (nd16) {\includegraphics[scale=0.1,width=0.7cm]
%                 {figures/khepera.pdf}};

% \node at (axis cs:90,130) (nd17) {\includegraphics[scale=0.1,width=0.7cm]
%                 {figures/khepera.pdf}};

% \node at (axis cs:90,112) (nd17) {\includegraphics[scale=0.1,width=0.7cm]
%                 {figures/khepera.pdf}};

\end{axis}
    \end{tikzpicture}
    \caption{Diagram of a typical binary collective decision-making scenario.}
\end{figure}
We investigate the effect of congestion on a binary decision-making problem
where the robots assess the quality of two sites via
sampling and collectively determine the superior location (see
fig.~\ref{fig:cartoon}). The robots share an arena of a given size (the ``space
medium''), a limited communication medium, have a collision prevention behavior,
and a belief propagation mechanism through local communication. We identify two
types of congestion: movement congestion, which happens when robots hinder each
other's movements and is proportional to the arena occupancy and the robot
behavior; and communication congestion, which is caused by belief propagation
conflicts that depend on the recency of the belief, communication range and the accuracy of the belief.

We answer three research questions:
\begin{compactenum}
\item What are the effects of movement and communication congestion
  w.r.t media occupancy?
\item What could be the essential factors that contribute to congestion?
\item Does introducing additional coordination mechanisms reduce congestion?
\end{compactenum} 

% We design experiments in a physics-based simulator~\cite{Pinci11} with an
% increasing number of robots (increasing media occupancy) and deploy three
% strategies for collective decision making: 1. Honey bee inspired, 2. Stigmergy
% based, and 3. Division of Labor. With the Honey Bee
% strategy~\cite{SeeleyHoneybee}, we observe how positive feedback from the agents
% induces movement congestion. The virtual stigmergy approach uses a replicated
% data structure shared among the robots to propagate the belief across the swarm:
% this approach lowers the effect of movement congestion while increasing
% communication conflicts compared to the Honey Bee. Finally, applying Division of
% Labor with clear robot roles (samplers sampling sites and networkers propagating
% belief) enables minimal convergence time, even with increased communication
% conflicts.

The remainder of the paper is organized into the following sections: we discuss
some related works in~\ref{sec:related_works}, explain the problem setting
in~\ref{sec:problem_setting}, explain the strategies mentioned above
in~\ref{sec:approach}, report the results~\ref{sec:results} and draw some
conclusions in~\ref{sec:conclusions}.

\section{Related work}\label{sec:related_works}
% We briefly review some of the works in collective decision-making, congestion
% prediction or mitigation, and Division of Labor literature used in multi-agent
% systems.

\textbf{\emph{Collective decision-making:}} There is a vast literature of self-organizing
discrete collective decision-making (DCDM) strategies inspired by the
house-hunting behavior~\cite{SeeleyHoneybee, Franks2002} and positive feedback
modulation~\cite{Garnier2007} from the waggle dance of honey bees, where the
task of the swarm is to find the best of two discrete options spatially
segregated into zones (see Figure~\ref{fig:cartoon}). Each agent assesses the qualities
of sites, advertises their opinions proportionally to the quality of
zone and applies a voter based ~\cite{ValentiniWeighter} or
majority~\cite{Valentini2016} based decision rule. This problem is extended to
dynamic site qualities in~\cite{Prasetyo2019}.

In a slightly different setting, the swarm is tasked to find the frequency of
features spread all over the environment for a single
feature~\cite{Valentiniperception}, with noise~\cite{chin2022minimalistic}, and
multiple features~\cite{Ebert2018MultiFeatureCD}. Further Bayesian approaches
were formulated and studied for static~\cite{Julia2017,shan2021}
and dynamic environments~\cite{PfiserKai2022}.

Continuous collective decision-making (CCDM), on the other hand, deals with
finding consensus on some environmental feature (e.g.,
intensity~\cite{RaoufiMohsen2021}, environmental edge~\cite{Khaluf2017} and tile
density~\cite{Khaluf2020}).
% Jamishdpey et al.~\cite{Jamshidpey2022} study the
% collective perception of object density in the surrounding environment using a
% self-organized hierarchy.

None of the above decision-making strategies address the movement and
communication congestion arising from increasing system size. In this work, we
adapt the existing static, discrete collective decision-making setting and
strategy from~\cite{ValentiniWeighter,Valentini2016} coupled with feature-like
distribution limited to the zones~\cite{Valentiniperception}, combining the Nest
site selection and collective perception from swarm robotics literature.

\textbf{\emph{Congestion prediction or mitigation:}} Ants~\cite{Couzin2002} and
humans~\cite{Helbing2001} form self-organizing lanes that help avoid congestion.
In artificial systems, some measures used to quantify movement congestion are
throughput and collisions. Throughput encodes the ability of multiple robots to
reach a given target, and Dos Passos et al.~\cite{dospassos2022} use throughput
to compare congestion of various strategies. Yu and Wold~\cite{Yu2023} deploy
ConvLSTMs to predict delays caused by congestion in a centralized warehouse
management system and increase throughput. Proximity encounters and collisions
are often used as a measure for congestion: A strategy to avoid head-on
collisions between two groups of swarms was proposed in~\cite{Marcolino2009},
and Wu et al.~\cite{Fang2019} propose collision-aware task assignment to
minimize congestion. Communication congestion is often correlated to a degraded
medium offering lower bandwidths~\cite{Barcis2021,Rybski2002}. In robot swarms,
propagating beliefs with an increasing number of robots can generate conflicts
on top of these bandwidth concerns. We use communication conflicts as a metric
to quantify communication congestion.

\textbf{\emph{Divison of labor:}} A taxonomy of heterogenous robot swarms
includes two high-level classes: behaviorally (software) and physically
(hardware) different swarm members~\cite{bettini2023heterogeneous}.
Behaviourally distinct swarm members often have uniform hardware with
role-specific behavior as in~\cite{Ferrante2015EvolutionOS}, where agents
specialize to become collectors or droppers in a food transporting task.
Behavioral variations can be dynamically triggered based on environmental
features~\cite{Ayanian2019DART} or could be static to divide tasks, as in
shepherding~\cite{Ozdemir2017ShepherdingWR}. Swarms of physically distinct
robots can benefit from traversing parts of the environment with aerial and
ground robots~\cite{varadharajan2020swarm} or collaboratively mapping the
environment with various sensors~\cite{lajoie2023swarmslam}. Having physically
and behaviorally different swarm members can offer efficient task completion
during a collaborative mapping task~\cite{Miller2022Stronger}. A variety of
missions have demonstrated the benefits of using physically and behaviorally heterogeneous swarms
in missions like search and retrieval task~\cite{dorigo2013novel} and formation
control~\cite{Saska2014CoordinationAN}. In this work, we use a physically
uniform and behaviorally distinct swarm to study decision-making in the
Division of Labor technique.

\begin{figure*}[htb]
    \tiny
    \centering
    \scalebox{.9}{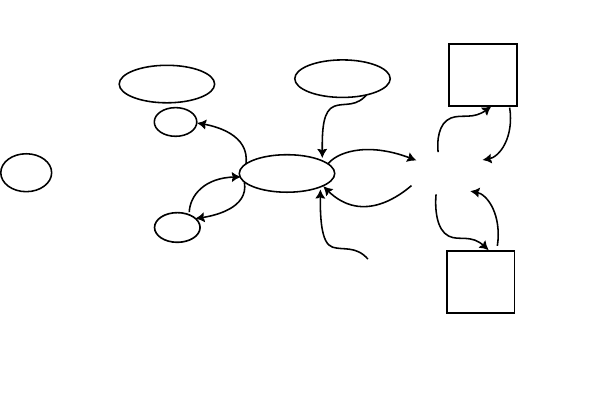}
    \scalebox{.9}{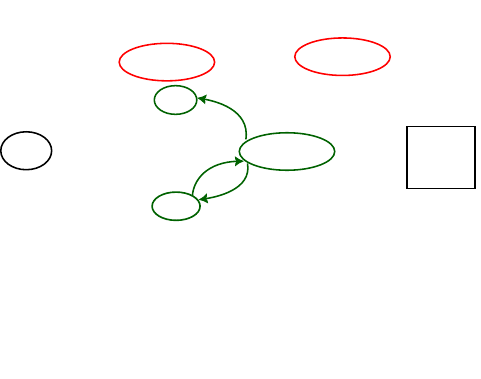}
    \caption{The state machines illustrate the behavioral states of robots during the three strategies Honey bee, Stigmergy, and Division of Labor SM(1-3). Every robot in the swarm deployed a corresponding state machine during the evaluation runs.}\label{fig:sms}
\end{figure*}
\section{Problem Setting}\label{sec:problem_setting}
We consider an arena of size $U \times V$ subdivided into three zones: A, B, and
Nest. Each sampling zone (A,B) is composed of a uniform distribution of a fill
ratio comprising of white and black tiles representing the quality of the site
$\rho \in [0,1]$, where $0$ represents complete black and $1$ represents
complete white. A swarm composed of $N$ Khepera IV robots (modeled as $\dot{x}_i
= u_i$, where $x_i \in \mathbb{R} ^2$ is the position of the robot, with a
circular communication model of range $R$, and with a ground footprint of 0.045
$m^2$), equipped with 4 ground ($G_i=\{G_i^0,..,G_i^3\}$), 8 proximity ($P_i =
\{P_i^0,..,P_i^7\}$), and 8 light sensors ($L_i = \{L_i^0,..,L_i^7\}$). Each
robot has to individually collect $S_T$ samples using the ground sensors,
calculate and communicate its belief state ($0 \leq bel_i \leq 1$), and avoid
collisions. The swarm collectively decides the highest quality zone (A or B).
There are five beacon robots placed at the boundary of both the sampling zones
that constantly broadcast zone option messages (i.e, A or B) to help robots
situate themselves inside the sampling zones. If a robot receives no broadcast
message it is considered to be in the Nest zone. To help robots move between the zones
there is a light placed above zone A, following the light gradient using the
light sensors ($PT$ - Phototaxis) leads the robots to zone A while doing the
opposite ($!PT$ - Antiphototaxis) leads the robots away from zone A to the zone
B.

\section{Approach}~\label{sec:approach}
We consider three state-machines outlined in fig.~\ref{fig:sms}: Honey Bee, Stigmergy, and Division of Labor decision-making strategies. These state machines are made of robot behaviors such as Diffusion ($DF$), Collision Avoidance ($CA$), Phototaxis ($PT$), and AntiPhototaxis ($!PT$).\\
\textbf{\emph{Collision Avoidance ($CA$):}} To avoid obstacles and other robots,
every robot uses the proximity sensors $P_i$. An obstacle vector is constructed
as $V_i^o = \frac{\sum_{i=0}^{7}P_i^i}{||P_i||} $ and applied as a control input
to the robot as shown below, where $S_o$ is a scaling factor and $O_{lt}$ and
$O_{at}$ are threshold parameters for obstacle avoidance.
\begin{align}
    \dot{x}_i &= \frac{-S_oV_i^o}{||V_i^o||}, & ||V_i^o|| \geq O_{lt} \ \& \ & -O_{at}  \geq \angle V_i^o \geq O_{at} 
\end{align}
This behavior moves the robot in the opposite direction of the aggregated obstacle vector ($V_i^o$), hence locally avoiding collisions.\\
\textbf{\emph{Phototaxis and AntiPhototaxis ($PT$ and $!PT$):}} To move between
zones robots use the light sensors $L_i$ whose readings are defined by the
equation $||L_i^i|| = (I/x)^2$, where $I$ is the reference intensity and $x$ the
distance between the light and the sensor. A Light vector is constructed as
$V_i^l=\frac{\sum_{i=0}^{7}L_i^i}{||L_i||}$ and applied as a control input to
the robot as shown below unless a collision is detected, where $S_l$ is a
scaling factor.
\begin{align}
    \dot{x}_i &=
    \begin{cases} 
    \frac{S_lV_i^l}{||V_i^l||}\ ,&PT \\
    \frac{-S_lV_i^l}{||V_i^l||}\ ,&!PT
    \end{cases} \label{equ:center_push}
\end{align}
\textbf{\emph{Diffusion ($DF$)}}: When the robot needs to explore the sampling
zones to collect samples or mix with other agents for efficient information
propagation while advertising the beliefs in the Nest zone, it uses diffusion,
where the robot just moves forward in the local frame with the maximum speed
($u_i^x = M_s$, $u_i^y=0$), unless a collision is detected. Collisions with
other robots and obstacles helps the robot diffuse.
 
\subsection{Honey Bee}
In this decision-making strategy, the robots are first initialized in a random
distribution in the Nest zone. Robots with even/odd IDs are assigned ($Z_i$) to
sample the zone (A/B) respectively. To reach the zone (A/B) the robots perform
$PT/!PT + CA$. When they reach the zone, they will receive a broadcast from the
A/B zone beacons. Upon reaching the zone, the robots diffuse ($DF+CA$) and start
collecting $S_T$ no of samples from their ground sensors, where each sample is
$bel_i(t) = \frac{\sum_{i=0}^{3}G_i^i}{||G_i||}$. After this, they come back to
the Nest zone to disseminate their averaged beliefs by executing the opposite
behavior $!PT/PT + CA$ used to reach the zones A/B. Upon reaching the Nest zone,
robots broadcast their averaged individual beliefs calculated as $avg_i^{bel} =
\frac{\sum_{t=1}^{S_T}bel_{i}(t)}{S_T}$ while diffusing ($DF+CA$) for a period
of time ($W_T \propto avg_i^{bel}$). This positive modulation of belief disemination is done to influence
more robots to choose the best site. Before the end of this period of time
($W_T$), robots start collecting their local neighbors ($n_i$) beliefs. Robots
further divide $n_i$ into two sets $nA/B:=\{j | j \in n_i\ \textit{and}\ Z_j =
A/B\}$. Along with their own beliefs, robots calculate two aggregated averages
one each for zone $Z_i$ and $!Z_i$
\begin{align}
    agg_{i}^{Z_i} = \frac{\sum_{j=1}^{|nZ_i|}avg_j^{bel} + avg_i^{bel}}{|nZ_i| + 1}\label{eqn:local_1}
\end{align}
\begin{align}
    agg_{i}^{!Z_i} &=
    \begin{cases}
        \frac{\sum_{j=1}^{|!Z_i|}avg_j^{bel}}{|n!Z_i|}\\
        0.0 & |n!Z_i|=0\label{eqn:local_2}
\end{cases}
\end{align}
if $agg_{i}^{!Z_i}$ $>$ $agg_{i}^{Z_i}$, $Z_i$ is updated to $!Z_i$ (positive modulation recruiting more robots towards the higher quality site, represented by blue lines in the left of fig. ~\ref{fig:sms}), otherwise it remains the same and the cycle is continued. The experiment is continued until all the robots form the same opinion. 

This differs from the approaches used in
the~\cite{ValentiniWeighter,Valentini2016} in two ways. The qualities of the
zone aren't directly broadcasted when the robots enter the zone, the robots
calculate them by using their ground sensors. This change was done to make
robots explore the zone, which is more realistic than the scenarios considered
in~\cite{ValentiniWeighter,Valentini2016} and this approach further emphasizes
the effect of movement flexibility in collective decision-making, as robots now
have to move within zones. The second change is that the individual averaged
beliefs (not opinions) are broadcasted, to have easier decision-making during
tie-breaks and have a belief consensus with virtual stigmergy. This change
requires very little communication overhead.
\subsection{Stigmergy}
In this decision-making strategy, we adopt the same state machine from the Honey Bee approach but instead of using a local communication broadcasts, we use a versioned local communication approach (virtual stigmergy~\cite{carlo2016}) to store the aggregated beliefs of both zones in separate entries ($agg^{A/B}$). Virtual stigmergy creates a shared tuple memory among the robots, where each entry contains a key identifier, Lamport clock (version number), robot id modifying the value and the value to be stored. Robots in the swarm are allowed to read and write to the local memory of the tuple value. Each access to the local memory creates a message to be broadcast in the local neighborhood. Whenever a robot receives a more recent update to the tuple, it updates the local memory and broadcasts the entry, allowing for more recent entries to be propagated.

The entries in the virtual stigmergy are synchronized as long the robots are connected~\cite{carlo2016}, i.e., a communication path exists between any two connected robots. With virtual stigmergy, robots can communicate with other robots even with movement congestion. With this property, it doesn't make sense for the robots to spend time advertising their averaged beliefs proportional to the average belief ($avg_{i}^{bel}$). Therefore $W_T$ is constant irrespective of the quality of the site. $W_T$ has to be still non-zero as mixing robots is still essential for synchronizing entries. At the beginning of this period ($W_T$) the robots read the entry of the zone they are assigned ($Z_{i}$) and update it using the equation~\ref{eqn:virtual_stig}.  
\begin{align}
    agg^{Z_{i}} &= agg^{Z_{i}} + w(avg_{i}^{bel} - agg^{Z_{i}}) \label{eqn:virtual_stig}  
\end{align}
where $w$ is the weight parameter. Instead of calculating the $agg^{A/B}$ like equations~\ref{eqn:local_1},~\ref{eqn:local_2}, the robots use the values from the stigmergy (Note that the subscript $i$ is dropped for $agg^{Z_{i}}$ in equation~\ref{eqn:virtual_stig}). As multiple robots might try to update the stigmergy at the same time (communication conflicts), a conflict resolution manager is used that keeps track of the maximum value for the aggregate belief for all robots. We count the number of conflicts occurring in this manager as the number of communication conflicts.

\subsection{Division of Labor}
It can be seen that every robot in Honey Bee approach pursues two roles: sampling and advertising, this mandates movement of robots between zones. In this approach instead, we assign fixed permanent roles for robots: samplers and networkers hence spatially segregating them into zones (A,B) and Nest respectively. Robots are randomly initialized in the Nest zone. One-third of robots are assigned ($Z_i$) to sample zone A, they follow the same state machine from the Honey Bee approach until they enter zone A. Similarly one-third of robots are assigned to be zone B samplers. This approach differs from the previous approaches after this point as the robots stay and diffuse in their zones (essentially disabling the positive modulation leading to the recruitment of more and more robots towards the higher quality zone in previous approaches) and after every $S_T$ number of samples collected, they read both the entries to keep track of the best zone opinion and update the aggregated belief in the stigmergy (similar to equation~\ref{eqn:virtual_stig}). The remaining one-third of robots stay and diffuse in the Nest zone acting as networkers by providing connectivity between the samplers for efficient belief propagation between both the sampling zones. Additionally, they also constantly keep track of the best zone opinion. This is continued until all the robots form one opinion. 

\section{Results}\label{sec:results}
\begin{figure*}[tbp]
    \centering
    \includegraphics[width=0.325\textwidth]{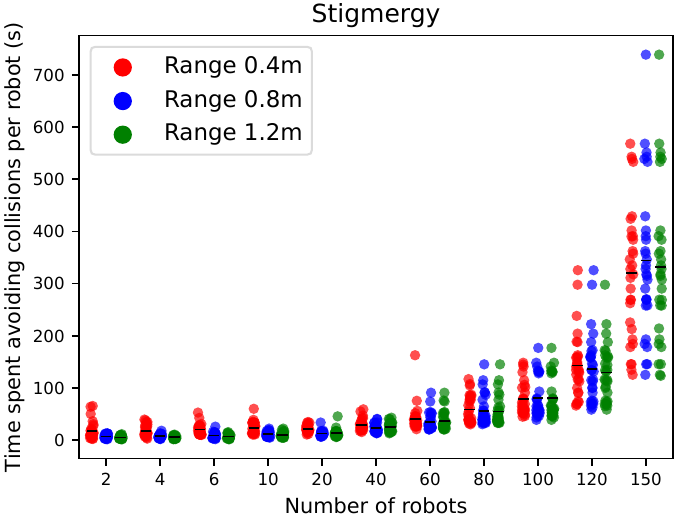}
    \includegraphics[width=0.325\textwidth]{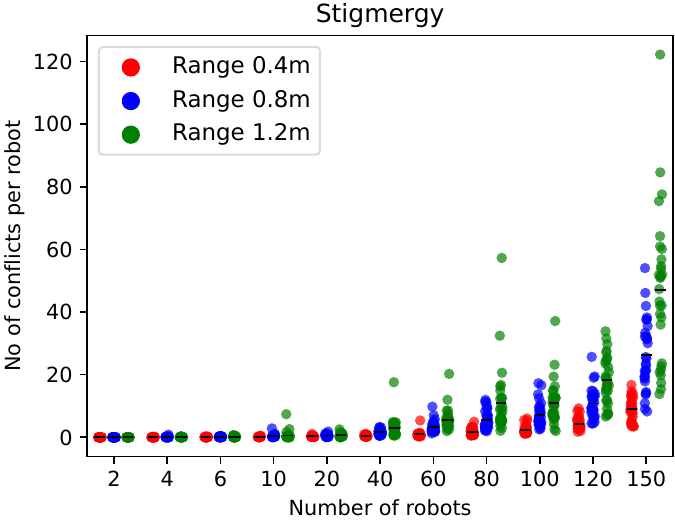}
    \includegraphics[width=0.325\textwidth]{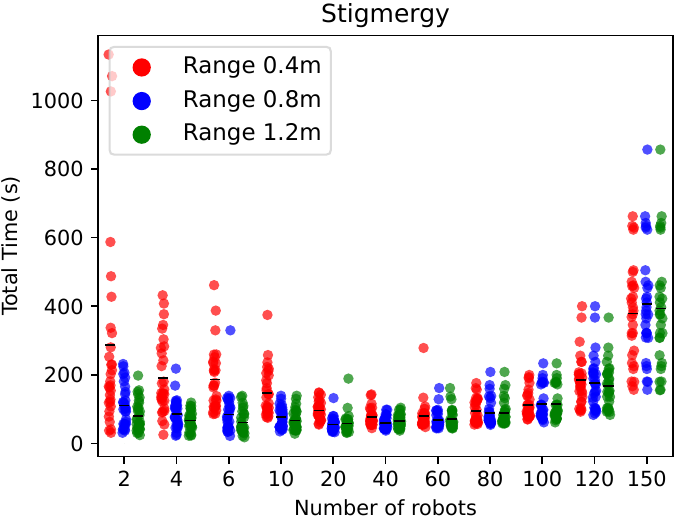}
    \includegraphics[width=0.325\textwidth]{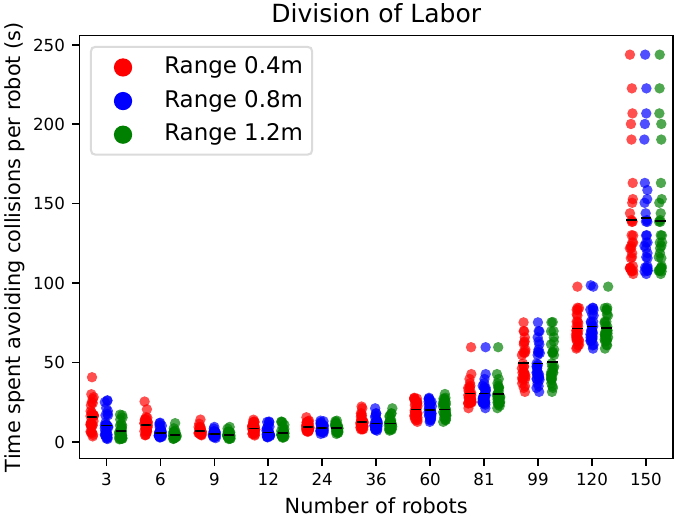}
    \includegraphics[width=0.32\textwidth]{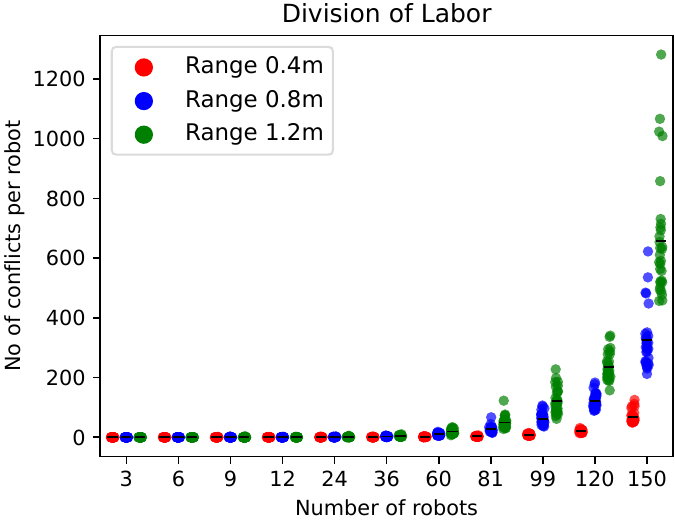}
    \includegraphics[width=0.32\textwidth]{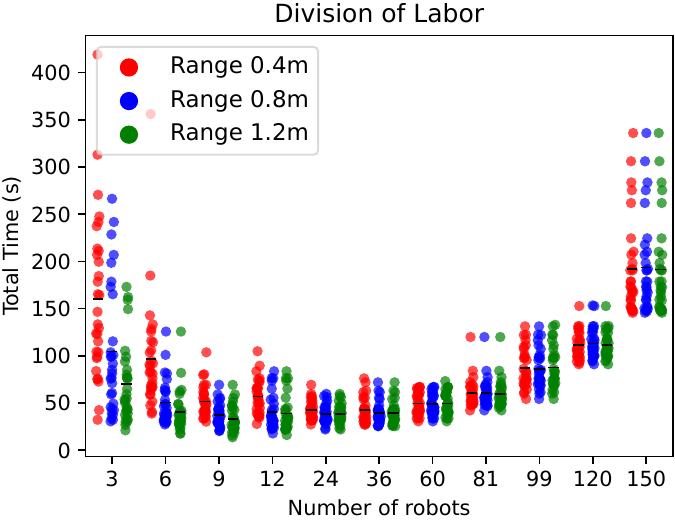}
    \includegraphics[width=0.325\textwidth]{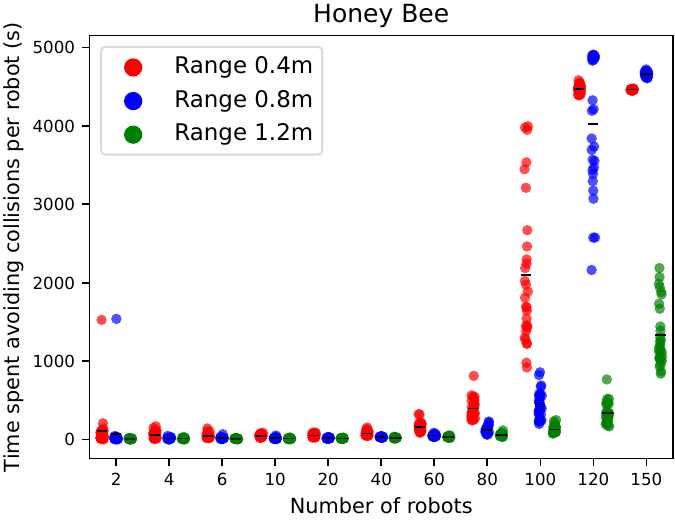}
    \includegraphics[width=0.325\textwidth]{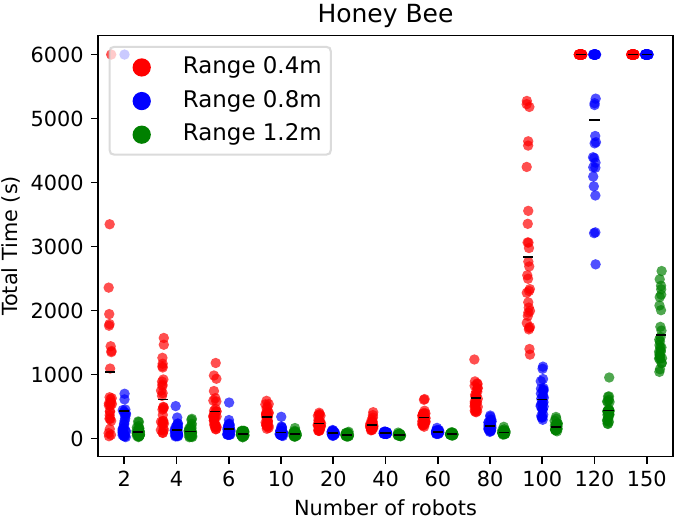}

    \caption{Congestion trends for scalability metrics for all three approaches.  It would appear that the total time trends for all three approaches have a similar pattern for any decentralized system~\cite{superlinear_heiko_2018} and the optimal number of robots for our setting would be around $(N\in\ \{20-60\})$ roughly. All experiments of Stigmergy and Division of Labor converged to the superior quality opinion before 6000 s timeout period, whereas some experiments of $(N=120,\ R=0.8),(N=2,\ R=\{0.4,0.8\})$ and all experiments in $(N=120,\ R=0.4),(N=150,\ R=\{0.4,0.8\})$ for Honey Bee approach failed to converge to any opinion.}
\label{fig:performance_metrics}
\end{figure*}
\begin{figure*}[tbp]
    \centering

    \includegraphics[width=0.190\textwidth]{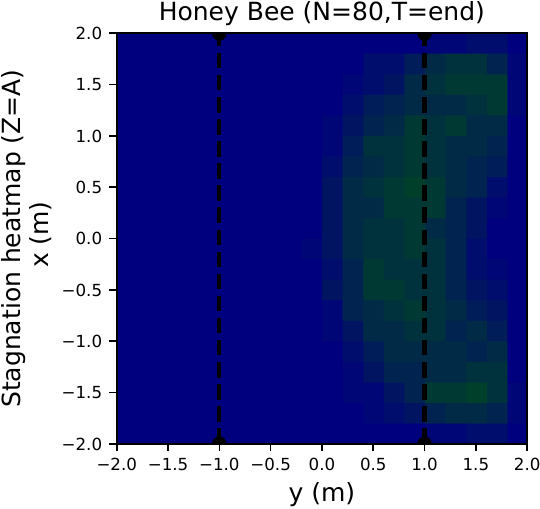}
    \includegraphics[width=0.180\textwidth]{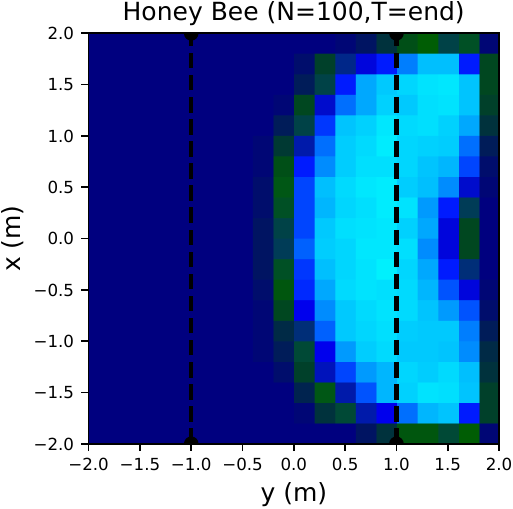}
    \includegraphics[width=0.180\textwidth]{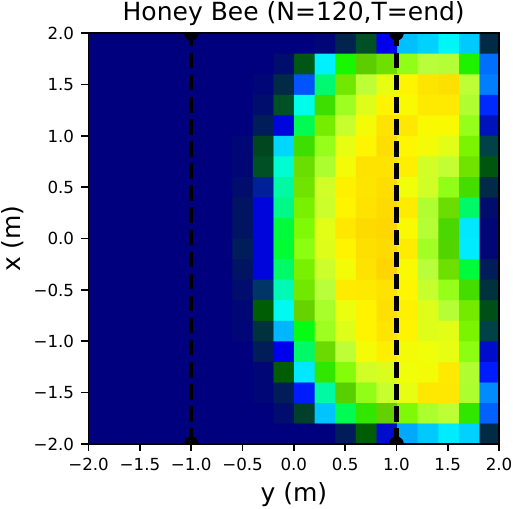}
    \includegraphics[width=0.205\textwidth]{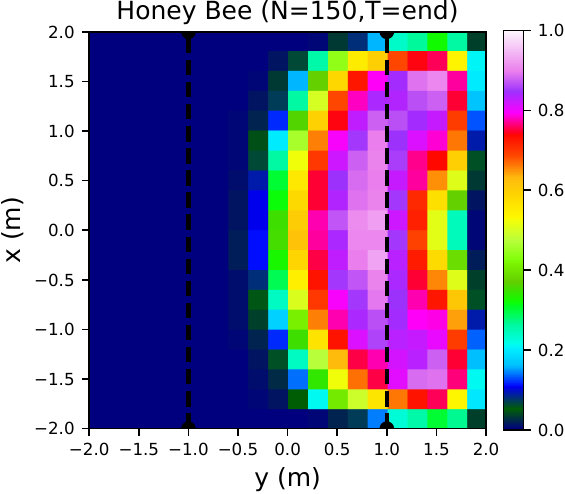}

    \includegraphics[width=0.190\textwidth]{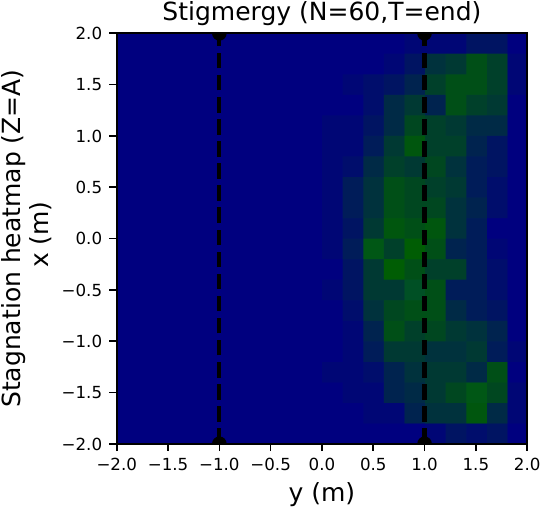}
    \includegraphics[width=0.180\textwidth]{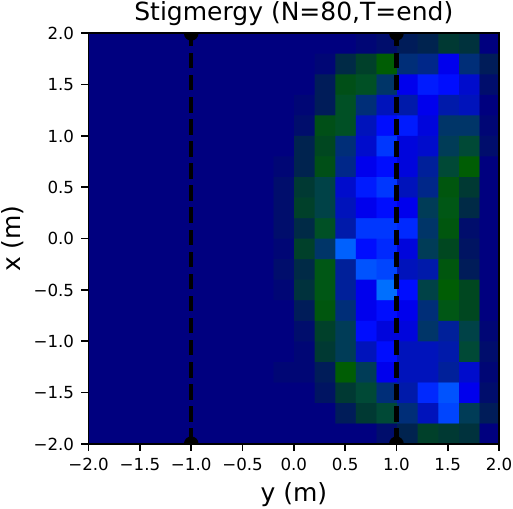}
    \includegraphics[width=0.180\textwidth]{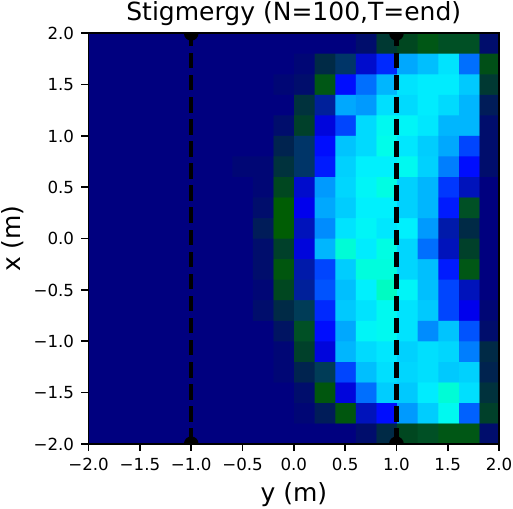}
    \includegraphics[width=0.180\textwidth]{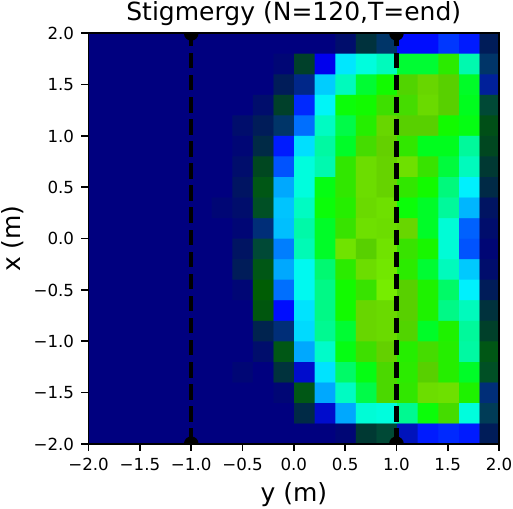}
    \includegraphics[width=0.205\textwidth]{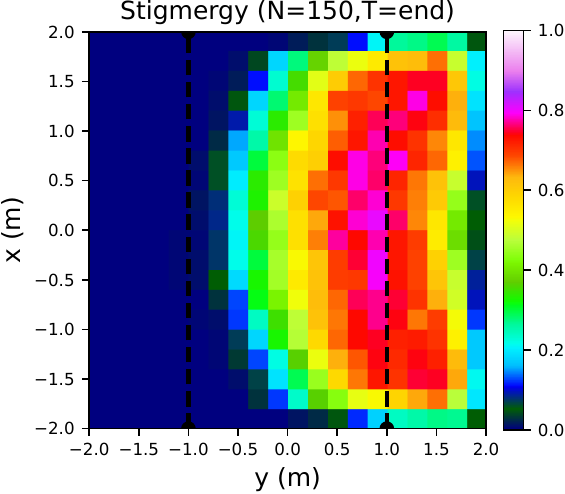}

    \caption{Stagnation heatmaps for the positive modulation approaches are shown in this figure for the combination $Z=A$, $R=0.4\ m$, $St_{T}$=1s (10 simulation timesteps). For the Honey bee approach $[T_s,T_f]$ = $[95\%,\ 100\%]$ and Stigmergy approach $[T_s,T_f]$ = $[85\%,\ 100\%]$. The barrier is of higher magnitude near the (zone A-Nest) boundary and has a decreasing radial gradient (in bands) away from zone A (the gradient follows a similar pattern to light intensity from the light centered at the end of zone A). Honey Bee approach is significantly more congested for a fixed combination compared to the Stigmergy approach (Honey Bee row is normalized by 5000 and Stigmergy row is normalized by 1000.)}

\label{fig:visitation_frequency_sdm_bdm}
\end{figure*}
\begin{figure}[tbp]
    
    \includegraphics[width=0.155\textwidth]{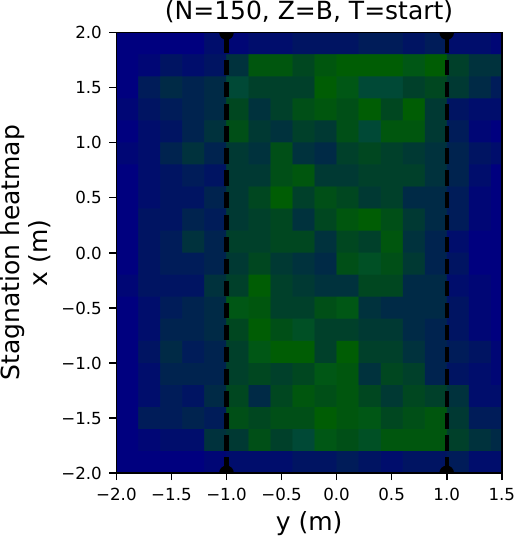}
    \includegraphics[width=0.152\textwidth]{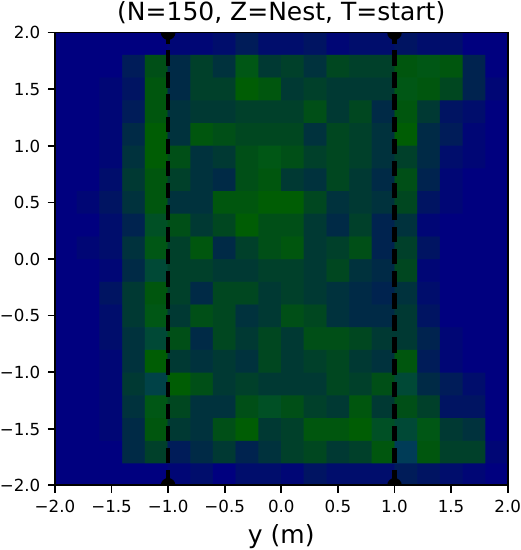}
    \includegraphics[width=0.155\textwidth]{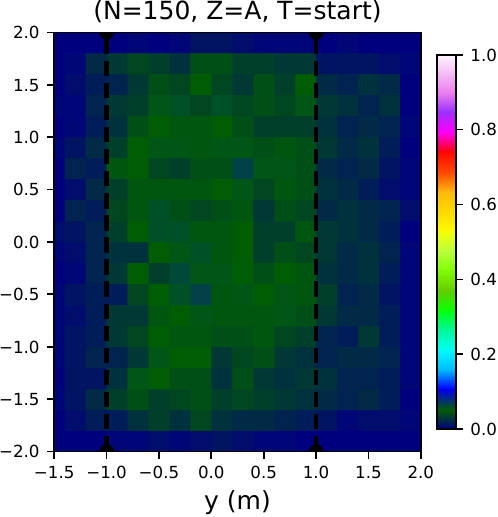}
    \includegraphics[width=0.155\textwidth]{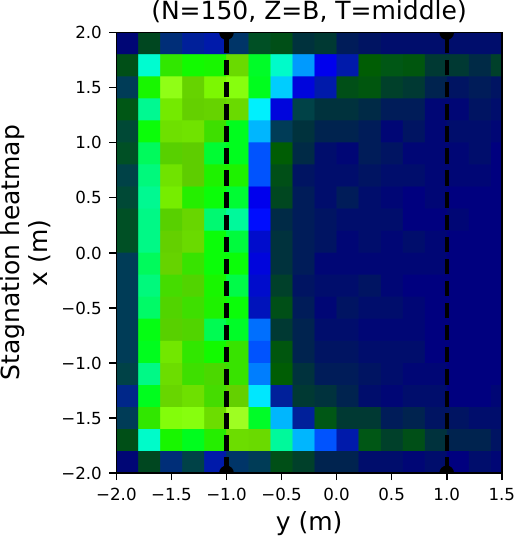}
    \includegraphics[width=0.152\textwidth]{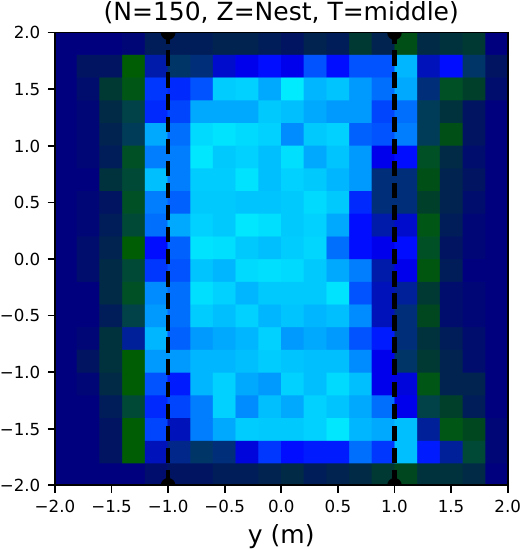}
    \includegraphics[width=0.155\textwidth]{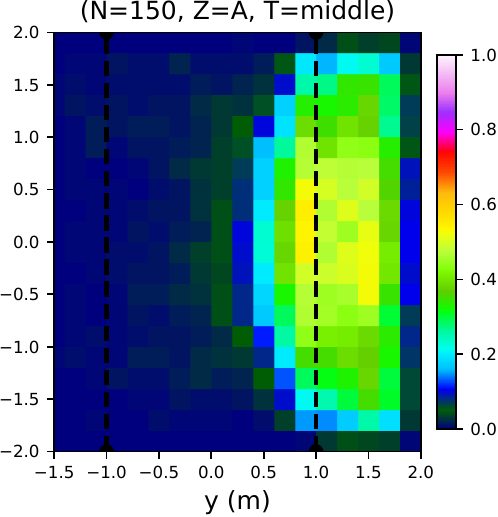}
    \includegraphics[width=0.155\textwidth]{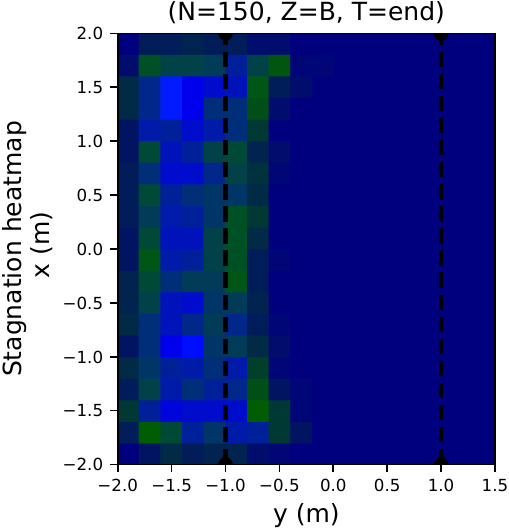}
    \includegraphics[width=0.152\textwidth]{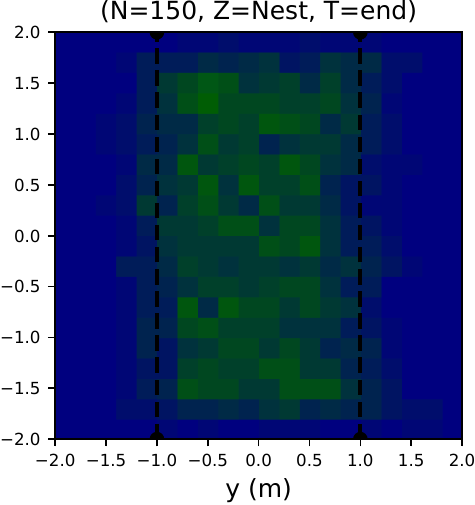}
    \includegraphics[width=0.155\textwidth]{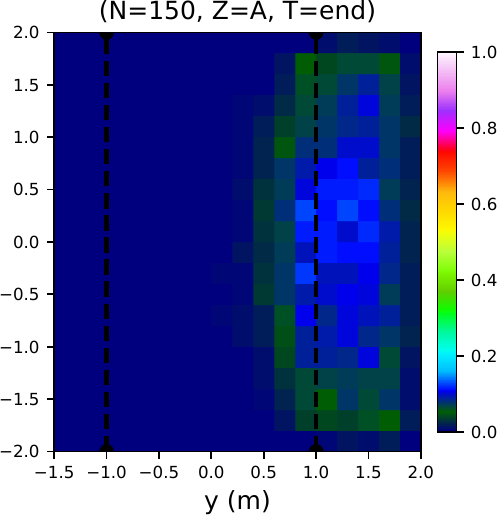}
    \caption{Stagnation heatmaps for Division of Labor approach is shown in this figure for $N=150$ $R=0.4\ m$, $St_{T}$=1s (10 simulation timesteps). T = start, represents $[T_s,T_f]$ = $[0\%,\ 15\%]$, T = middle, represents $[T_s,T_f]$ = $[15\%,\ 85\%]$, and T = end, represents $[T_s,T_f]$ = $[85\%,\ 100\%]$. It can be seen that the magnitudes of stagnations are lesser compared to fig~\ref{fig:visitation_frequency_sdm_bdm} and stagnations outside the assigned zones occur only in the starting phases of experiments. (All the rows are normalized by 1000.)}

\label{fig:Visitation_frequency_hetero}
\end{figure}%
We investigate the scalability of the three approaches using the following metrics: 1. average time spent by every robot avoiding collisions with other robots and obstacles (arena walls), 2. average communication conflicts per robot while updating the virtual stigmergy, and 3. total time taken for all the robots to converge to highest quality opinion.

During all the experimental evaluations, we deploy the robots in a fixed arena dimension of $U=4\ m,\ V=4\ m$ and a Nest of size $2m\times4m$ with site A quality $\rho_{A} = 0.9$ and site B quality $\rho_{B} = 0.1$. We varied the number of robots $N \in \{2,\ 4,\ 6,\ 10,\ 20,\ 40,\ 60,\ 80,\ 100,\ 120,\ 150\}$ corresponding to a Nest robot density of $\{1.3,\ 2.6,\ 3.9,\ 6.5,\ 13.1,\ 26.2,\ 39.4,\ 52.5,\ 65.7,\ 78.8,\ 98.5\}$ $\times$ $10^{-2}$ for Honey Bee and Stigmergy based decision-making strategies. Similarly, for Division of Labor technique, varied the robot numbers $N \in \{3,\ 6,\ 9,\ 12,\ 24,\ 36,\ 60,\ 81,\ 99,\ 120,\ 150\}$ corresponding to a Nest robot density of $\{1.9,\ 3.9,\ 5.9,\ 7.8,\ 15.7,\ 23.6,\  39.4,\ 53.2,\ 65,\ 78.8,\ 98.5\}$ $\times$ $10^{-2}$. We set the communication range for all three techniques to $R \in \{0.4\ m,\ 0.8\ m,\ 1.2\ m\}$ and repeated each configuration 30 times with randomized robot placement following a normal distribution in the Nest zone.

\begin{figure}[tbp]
    \includegraphics[width=0.160\textwidth]{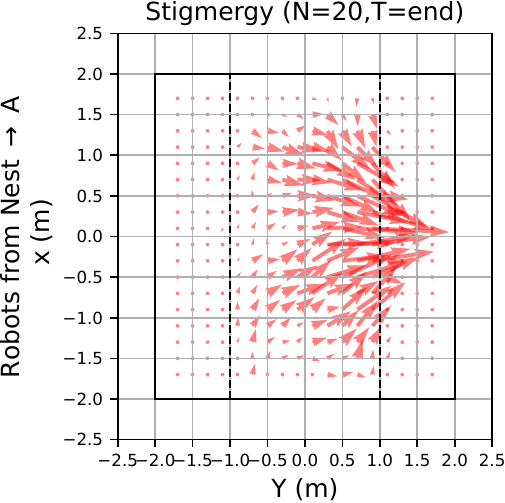}
    \includegraphics[width=0.152\textwidth]{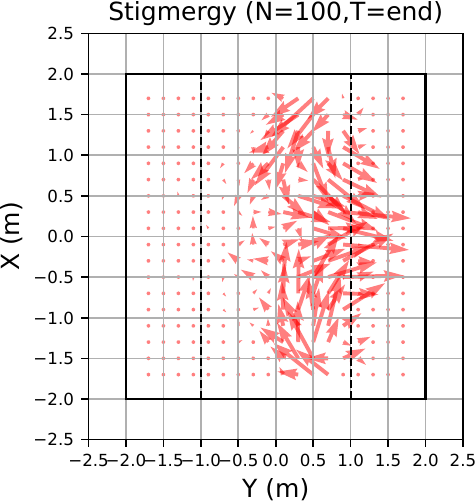}
    \includegraphics[width=0.152\textwidth]{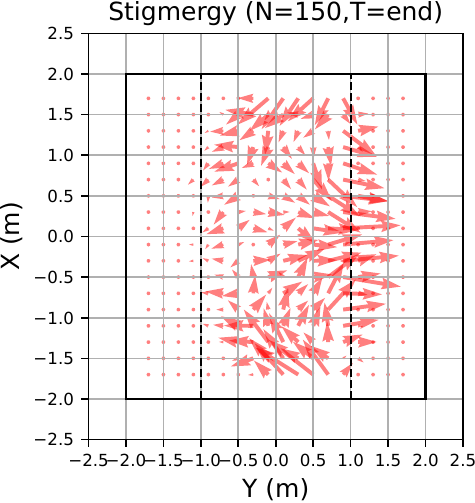}

    \includegraphics[width=0.156\textwidth]{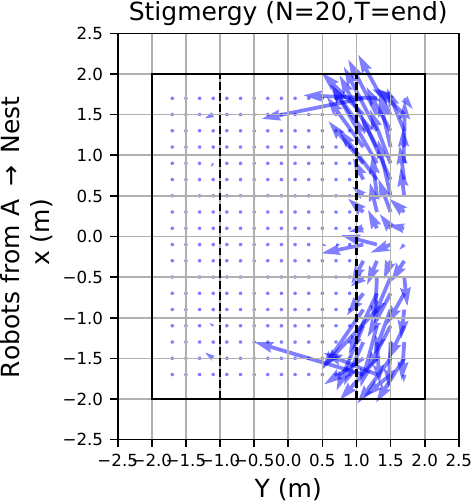}
    \includegraphics[width=0.154\textwidth]{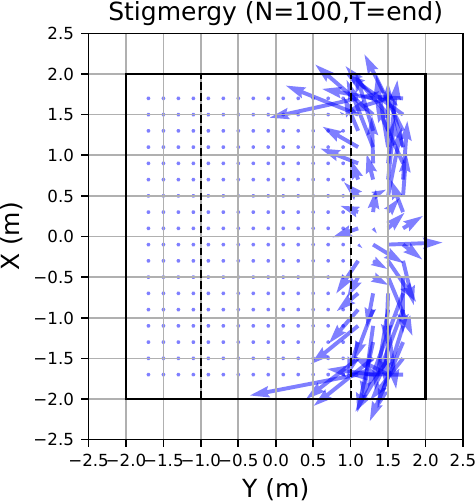}
    \includegraphics[width=0.154\textwidth]{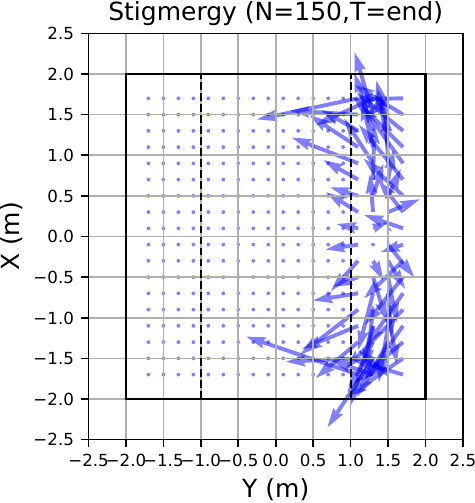}

    \caption{Movement changes for the combination $N\in \{20,\ 100,\ 150\}$, $R=0.4\ m$, $Z=A$, $[T_s,T_f]$ = $[85\%,\ 100\%]$. The top row shows the averaged movement change of robots entering Zone A from Nest
    to sample (Zone A followers in Stigmergy approach fig.~\ref{fig:sms}) and the bottom row shows the averaged movement change of robots entering Nest from Zone A (Nest followers in Stigmergy approach fig.~\ref{fig:sms}.)}
\label{fig:directional_change_stig}
\end{figure}

To further understand the effects of the movement congestion, we plot the accumulated stagnation heatmap (defined as a robot spending over $St_{T}$ seconds in grids of size ($0.2\times0.2$)) for an interval $[T_s,T_f]$ in fig.~\ref{fig:visitation_frequency_sdm_bdm} and fig.~\ref{fig:Visitation_frequency_hetero}. The averaged movement change gridmap divides the arena into grids of size ($0.2\times0.2$) for an interval $[T_s,T_f]$ in fig.~\ref{fig:directional_change_stig}. Averaged movement change for each grid cell is calculated by averaging the movement vectors of robots in the grid over consecutive time steps ($x_i(t+1) - x_i(t)$). The stagnation heatmap and movement change gridmap are averaged over all 30 repetitions of a given configuration. The stagnation heatmap shows the congestion in space, while the averaged movement change gridmap shows the movement of robots.

Leveraging the metrics introduced above, we make the following inferences:

(1) \textbf{\emph{Versioned local communication helps to overcome movement congestion, but doesn't decongest the system.}}
In fig.~\ref{fig:performance_metrics}, the convergence time plot from the first two rows show that the Stigmergy and Division of Labor strategy converge significantly faster than the Honey Bee inspired, despite a comparable stagnation pattern in Stigmergy based (see fig.~\ref{fig:visitation_frequency_sdm_bdm}). The presence of stagnation with the Stigmergy approach indicates that congestion still exists. Faster convergence correlates with the lower time spent avoiding obstacles.

(2) \textbf{\emph{Positive modulation in robot swarms increases the impact of stagnation and convergence time.}} 
A stagnation barrier (ref fig.~\ref{fig:visitation_frequency_sdm_bdm}) of increasing thickness with an increased number of robots occurs near the superior quality zone for Stigmergy and Honey Bee approaches. The barrier could result from positive modulation recruiting more and more robots to visit the higher quality zone as indicated in~\cite{ValentiniWeighter}. The barrier formation can also be inferred in fig.~\ref{fig:directional_change_stig}, where the averaged movement vectors of robots point towards each other, implying the robots' intention to move towards each other. The formation of a barrier significantly hinders the information propagation in the Honey Bee approach, where belief propagation occurs through local broadcasts and the ability of robots to move to exchange beliefs effectively. The effect of the stagnation barrier is more pertinent for larger robot density and smaller communication range for Honey Bee inspired.

(3) \textbf{\emph{Introducing structure through division of labor helps to minimize movement congestion.}} 
Fig.~\ref{fig:Visitation_frequency_hetero} shows the stagnation heatmap for the Division of Labor approach. The zone samplers and nest zone networker robots experience minimal stagnation within their respective zones as they are contained within their zones (except T=start, as robots are deployed in the Nest zone). The minimal stagnation in the grids directly reflects on the convergence time in fig.~\ref{fig:performance_metrics}, where convergence time and time spent on collisions are minimal compared to the other two approaches. However, communication conflicts are larger than in the Stigmergy approach as more updates to the robot beliefs propagate through the swarm. The Stigmergy approach still suffers from movement congestion, thereby influencing the ability of the robots to move, sample, and update the stigmergy, which results in fewer conflicts.

(4) \textbf{\emph{Longer communication ranges make a positive difference only with the local broadcast approach and make a negative impact with the versioned local communication strategy for a larger number of robots.}} Longer communication range combinations used in Honey Bee approach improve the total time and time spent avoiding collisions (ref fig.~\ref{fig:performance_metrics}) compared to shorter communciation ranges ($R=0.4\ m < 0.8\ m < 1.2\ m$) for any number of robots in the system except ($N=150,\ R=\ 0.8\ m$, which has a slight increase in the time spent avoiding collisions per robots compared to $N=150,\ R=\ 0.4\ m$). Whereas the number of conflicts arising with the versioned local communication approach increases with a longer communication range and a higher number of robots ($N>60$). As the movement congestion doesn't impact the propagation of beliefs with the versioned local communication approach, there is no significant improvement in the total time taken and time spent avoiding collisions for both these approaches compared to shorter communication ranges for a fixed number of robot combination (for $N>20$, ref fig~\ref{fig:performance_metrics}).

\section{Conclusions}~\label{sec:conclusions}

Current collective decision-making strategies rarely address congestion-related
issues. This will have huge implications when it comes to deploying robot swarm
systems in real-world scenarios, as these systems will scale poorly. In this
paper, we discuss the impact of movement congestion and belief propagation
conflicts on swarm behaviors, specifically collective decision-making. We find
that using versioned local communication and Division of Labor mechanisms helps
to reduce the impact of movement congestion, despite the increasing trends for
communication conflicts. Further research could look into congestion-aware
initialization strategies, congestion-aware collision avoidance, and dynamic
approaches to switch between different state machines for collective
decision-making systems. We believe our results transfer to other areas of swarm
robotics such as foraging, task allocation, collective construction etc. and
would welcome additional studies in these domains.

\bibliographystyle{IEEEtran}
\bibliography{ref}

\end{document}